\documentclass[10pt,twocolumn,letterpaper]{article}

\usepackage{cvpr}              

\usepackage[accsupp]{axessibility}  
\usepackage{graphicx}
\usepackage{amsmath}
\usepackage{amssymb}
\usepackage{booktabs}

\usepackage{multirow}
\usepackage[dvipsnames]{xcolor}
\usepackage{colortbl}
\usepackage{tipa}
\definecolor{mygray}{gray}{0.93}

\usepackage[pagebackref,breaklinks,colorlinks]{hyperref}

\def\confName{CVPR}
\def\confYear{2022}

\begin{document}

\title{Temporally Efficient Vision Transformer for Video Instance Segmentation}

\author{Shusheng Yang$^{1,3}$\thanks{This work was done while Shusheng Yang and Yu Li were at Applied Research Center (ARC), Tencent PCG.}, \ \  Xinggang Wang$^{1}$\thanks{Corresponding author, E-mail: {\tt xgwang@hust.edu.cn}.}, \ \ Yu Li$^{4*}$, \ \ Yuxin Fang$^{1}$, \\
Jiemin Fang$^{2,1}$, \ \ Wenyu Liu$^{1}$, \ \ Xun Zhao$^{3}$, \ \ Ying Shan$^{3}$ \\
\normalsize 
$^1$School of EIC, Huazhong University of Science \& Technology \\
\normalsize 
$^2$Institute of Artificial Intelligence, Huazhong University of Science \& Technology \\
\normalsize 
$^3$Applied Research Center (ARC), Tencent PCG \ \ $^4$International Digital Economy Academy (IDEA) \\
}
\maketitle

\begin{abstract}
Recently vision transformer has achieved tremendous success on image-level visual recognition tasks. To effectively and efficiently model the crucial temporal information within a video clip, we propose a \textbf{T}emporally \textbf{E}fficient \textbf{Vi}sion \textbf{T}ransformer (TeViT) for video instance segmentation (VIS). Different from previous transformer-based VIS methods, TeViT is nearly convolution-free, which contains a transformer backbone and a query-based video instance segmentation head. In the backbone stage, we propose a nearly parameter-free messenger shift mechanism for early temporal context fusion. In the head stages, we propose a parameter-shared spatiotemporal query interaction mechanism to build the one-to-one correspondence between video instances and queries. Thus, TeViT fully utilizes both frame-level and instance-level temporal context information and obtains strong temporal modeling capacity with negligible extra computational cost. On three widely adopted VIS benchmarks, \ie, YouTube-VIS-$2019$, YouTube-VIS-$2021$, and OVIS, TeViT obtains state-of-the-art results and maintains high inference speed, \eg, $46.6$ AP with $68.9$ FPS on YouTube-VIS-$2019$. Code is available at \url{https://github.com/hustvl/TeViT}.
\end{abstract}

\section{Introduction}

Video Instance Segmentation (VIS)~\cite{vis} is a representative and challenging video understanding task that requires detecting, segmenting and tracking video instances across frames simultaneously. Similar to other instance-level video recognition tasks, making full use of temporal context information is critical for building high-performance VIS systems. Vision transformer (ViT) \cite{vit}, which is based on self-attention~\cite{transformer}, has shown strong long-range context modeling ability and obtained great successes on image classification~\cite{vit, deit, halonet, swintransformer, pvt, msgtransformer, shuffletransformer}, object detection~\cite{detr, defdetr, conddetr, sparsercnn}, semantic segmentation~\cite{segformer, segmenter, maskformer}, instance segmentation~\cite{queryinst, solq, knet}, and video recognition~\cite{timesformer, videoswin, tokshift, vivit, vidtr, mvit, vtn}.

Recently, how to design ViTs for instance-level video understanding, especially VIS, becomes an emerging problem. Different from the detection transformers \cite{detr,defdetr,conddetr,sparsercnn, yolos}, semantic segmentation transformers~\cite{segformer, segmenter, maskformer}, and instance segmentation transformers \cite{queryinst, solq, knet}, which focus on $2$D contextual information modeling, VIS transformers additionally require to perform temporal context modeling. To this end, VisTR~\cite{vistr} firstly proposes a transformer encoder to fuse patch features from a sequence of frames using a CNN backbone and leverages a query-based decoder to predict video instances, IFC~\cite{ifc} introduces memory tokens to store frame-level features and performs cross-frame feature interaction by computing self-attention among memory tokens, and then decodes instance-level results using a conditional mask head.

In this paper, we focus on the efficiency of modeling temporal information for ViT-based VIS. This is a very important problem since (1) computing self-attention among all video patches has extremely high time and space complexity \cite{vistr}, (2) additional multi-head self attention ($\mathrm{MHSA}$) layers for temporal modeling have extra parameters and are sensitive to pre-training \cite{ifc}, (3) the CNN or transformer backbones in these methods \cite{vistr, ifc, querytrack, tcis} only support single frame feature extraction and fail to capture temporal information in the backbone stage. To remedy the above issues, we present Temporally Efficient ViT (TeViT) to fully utilize temporal contextual information for efficient and effective video instance segmentation.

TeViT contains a transformer backbone and a series of query-based VIS heads. In the backbone stage, we use messenger tokens~\cite{msgtransformer} to extract intra-frame information via self-attention and propose a messenger shift mechanism for frame-level context modeling, in which messenger tokens are divided into several groups to perform temporal shift with various of time steps. Different from previous VIS methods, the messenger shift transformer enables early temporal feature fusion. 
In the head stages, we convert the QueryInst~\cite{queryinst} instance segmentation head into our VIS head by reusing the multi-head self-attention ($\mathrm{MHSA}$)~\cite{transformer} parameters for instance-level temporal information interaction. The instance-level $\mathrm{MHSA}$ fuses the features for a single video instance among input frames, thus it realizes the concept of a video instance as a query.

Experiments are conducted on three large-scale VIS datasets, \ie, YouTube-VIS-$2019$~\cite{vis}, YouTube-VIS-$2021$~\cite{vis2021}, and OVIS~\cite{ovis}. New state-of-the-art (SoTA) performance has been obtained, \eg, TeViT obtains $46.6$ AP with $68.9$ FPS on YouTube-VIS-$2019$. Our main contributions are summarized as follows.

\begin{itemize}
\item TeViT is the first video instance segmentation transformer that can efficiently capture temporal contextual information at both frame level and instance level.
\item Benefiting from the flexibility of self-attention, the proposed temporal modeling modules, \ie, messenger token shift and spatiotemporal query interaction, both are friendly to the image-level pre-trained models, cost marginal extra computation overhead and parameters.
\item TeViT is a nearly convolution-free framework and obtains SoTA VIS results. In TeViT, the concepts of ``early temporal feature fusion" and ``a video instance as a query" shield lights on how to build effective video transformers for instance-level recognition tasks.
\end{itemize}

\section{Related Work}

\noindent\textbf{Video instance segmentation.}
How to achieve efficiently temporal modeling is always the focus of video tasks, such as video object segmentation (VOS) \cite{uvos, uvos, stm}, multi-object tracking and segmentation (MOTS) \cite{mots} and VIS~\cite{vis}. Though VOS and MOTS are very related with VIS, MOTS mainly focuses on the urban scene understanding and VOS aims at tracking specific object by a given mask. Representative VIS works are reviewed as follows. MaskTrack R-CNN~\cite{vis} extends Faster R-CNN~\cite{fasterrcnn} and Mask R-CNN~\cite{maskrcnn} to VIS with a tracking branch and external memory that saves instance features across multiple frames. MaskProp~\cite{maskprop} builds on the Hybrid Task Cascade Network~\cite{htc} and propagates instance region features to adjacent frames to perform temporal modeling. STEm-Seg~\cite{stemseg} treats video clip as $3$D spatiotemporal volume and captures temporal information by $3$D convolutional backbone network. CompFeat~\cite{compfeat} refines temporal features at both frame-level and instance-level. CrossVIS~\cite{crossvis} introduces a crossover learning scheme upon~\cite{fcos, condinst} to make use of contextual information across video frames. SeqMask R-CNN~\cite{seqmaskrcnn} establishes temporal relation across frames by adding an extra sequence propagation head upon Mask R-CNN. Both VisRGNN~\cite{visrgnn} and VisSTG~\cite{visstg} model temporal information in VIS by a graph neural network. VisTR~\cite{vistr} proposes the first fully end-to-end VIS method upon DETR~\cite{detr}, temporal contexts are fused by the multi-head attention mechanism in transformer encoder layers. IFC~\cite{ifc} presents inter-frame communication to exchange frame-level information. In this paper, we present a temporally efficient framework to model temporal contexts at both frame-level and instance-level.

\noindent\textbf{Vision Transformer.}
Transformer~\cite{transformer} is firstly proposed to model long-range sequence data in natural language process (NLP). ViT~\cite{vit} firstly adopts transformer to image domain. After that various high-performance vision transformers \cite{swintransformer, halonet, deit, pvt, pvtv2, msgtransformer, shuffletransformer} have been proposed as backbones for image understanding. Beyond serving as backbone networks, transformer has motivated lots of novel object detection \cite{detr, defdetr, sparsercnn, conddetr, yolos}, instance segmentation \cite{queryinst, solq, knet}, and semantic segmentation \cite{segformer, segmenter, maskformer} frameworks. Recently, VisTR~\cite{vistr}, IFC~\cite{ifc}, QueryTrack~\cite{querytrack}, and TCIS~\cite{tcis} bring transformer to video instance segmentation and achieve excellent performance. In this paper, we investigate how to efficiently model temporal context across video frames and propose TeViT. TeViT is a nearly convolution-free transformer while VisTR and IFC both use ResNet~\cite{resnet} backbone.

\noindent\textbf{Temporal context modeling.}
Temporal context modeling is the key issue in video understanding. A lot of works \cite{c3d, i3d, p3d, s3d, r213d} model temporal context by $3$D convolutional block. TSM~\cite{tsm} proposes an efficient temporal shift module by moving the convolutional feature map along the temporal dimension. Non-local network~\cite{nonlocal} applies self-attention to capture long-range spatiotemporal dependencies directly. Recently, TimeSformer~\cite{timesformer}, ViViT~\cite{vivit}, VidTR~\cite{vidtr}, and MViT~\cite{mvit} extend ViT to capture spatiotemporal context for video classification. Video Swin Transformer~\cite{videoswin} extends Swin Transformer~\cite{swintransformer} to video by conducting shift window $\mathrm{MHSA}$ in both space and time. TokShift~\cite{tokshift} proposes a temporal shift mechanism on $\mathtt{CLASS}$ tokens of ViT. Different from these video transformers focus on video classification, we target at building temporally efficient transformer for instance-level video understanding.

\begin{figure*}
    \centering
    \includegraphics[width=0.75\linewidth]{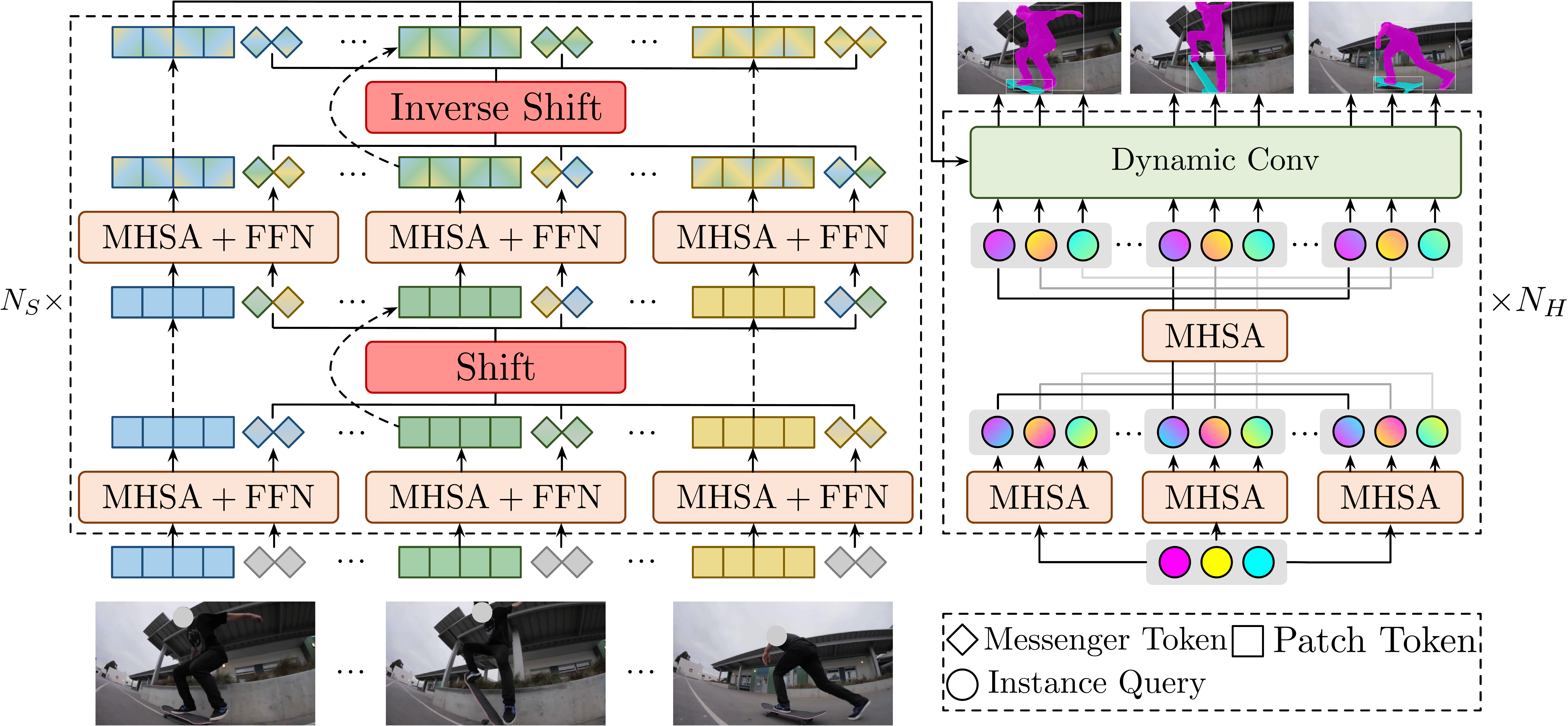}
    \caption{The overall illustration of our TeViT framework. TeViT contains a messenger shift transformer backbone and a series of spatiotemporal query-driven instance heads. The messenger shift mechanism performs efficient frame-level temporal modeling by simply shifting messenger tokens along the temporal axis. Spatiotemporal query interaction conducts two successive and parameter-shared multi-head self attention ($\mathrm{MHSA}$) with feed forward network ($\mathrm{FFN}$) upon video instance queries. The ``Dynamic Conv" design follows QueryInst~\cite{queryinst}. Best viewed in color.}
    \label{fig:overall_arch}
\end{figure*}

\section{Method}

\subsection{Overall Architecture}

The overall architecture of our VIS method TeViT is shown in Fig.~\ref{fig:overall_arch}, which contains a transformer-based backbone network and a query-driven head network. Given a sequence of video frames, the transformer backbone performs feature extraction and generates multi-scale pyramid features. The query-driven head network takes randomly initialized instance queries with backbone feature maps to predict video instances. Our whole network is end-to-end for both training and inference.

\subsection{Messenger Shift Transformer Backbone}
\label{sec:messenger_shift_transformer_backbone}

\begin{figure}
    \centering
    \includegraphics[width=1.0\columnwidth]{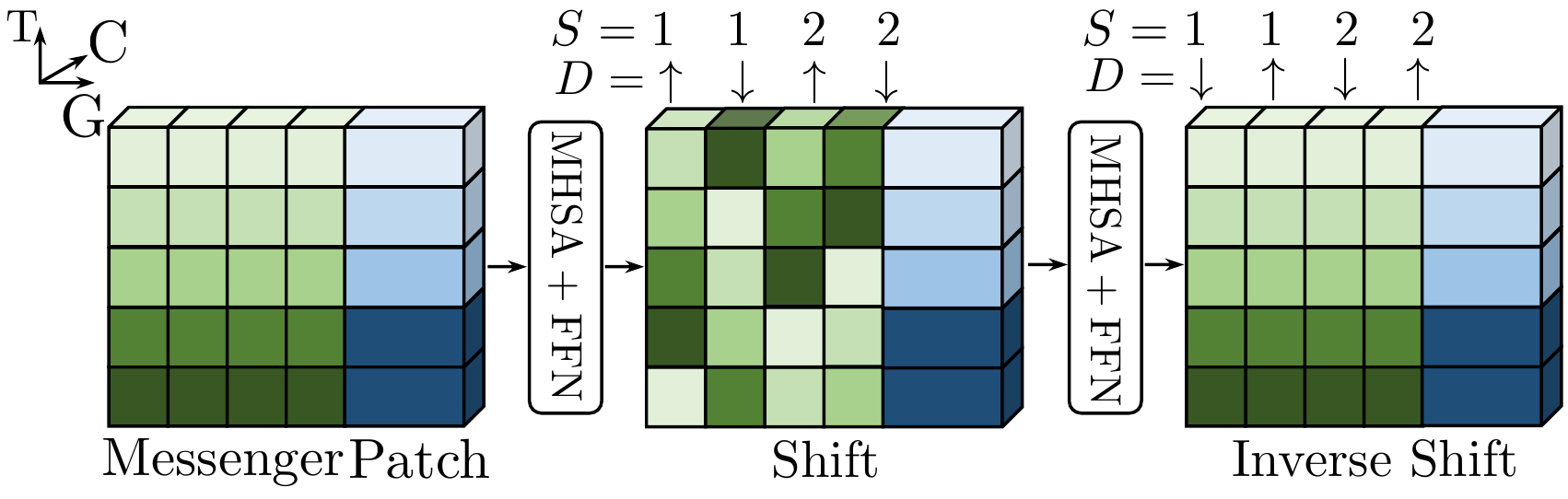}
    \caption{An illustration of the messenger shift mechanism. Messenger tokens are first segmented into several groups ($4$ in the figure) and then shifted along the temporal axis with different stride ($S$) and direction ($D$) to exchange frame-level information. For every two successive messenger shift mechanisms, we apply an inverse shift operation. As shown in figure above, the shift direction ($D$) of each token group in the second shift mechanism (right) is exactly inverse to the first (medium). Messenger tokens are shown by green cubes, while blue cubes denote the patch tokens.}
    \label{fig:messenger_shift}
\end{figure}

In previous VIS methods, the backbone networks only perform feature extraction in per-frame fashion \cite{vistr, ifc} and neglect the rich contextual information inherent in video frames. In contrast, inspired by MSG-Transformer \cite{msgtransformer}, we propose messenger shift transformer (MsgShifT) which performs highly efficient temporal context modeling in a bottom-up manner, as shown in Fig.~\ref{fig:overall_arch} (left). Without loss of generality, we build MsgShifT based on the pyramid vision transformer (PVT)~\cite{pvt, pvtv2}.

To be specific, given an input video with $T$ frames of resolution $H \times W$, denoted as $\left\{x_{i}\right\}_{i=1}^{T} \in \mathbb{R}^{T \times 3 \times H \times W}$, we first divide these frames into $T \times \frac{HW}{P^2}$ patch tokens frame-by-frame, where $P$ denotes the size of each patch. Then we feed the flattened patch tokens to a linear projection and get embedded patches $\left\{f_i^0\right\}_{i=1}^T$ with size of $T \times \frac{HW}{P^2} \times C$, $C$ denotes the channel dimension.
Meanwhile, a group of randomly initialized learnable embeddings with size of $M \times C$ are introduced as messenger tokens, denoted as $m^0$, where $M$ indicates the number of messenger tokens.
Then we simply copy and concatenate messenger tokens with patch tokens:
\begin{equation}
    \left\{[f_i^0, m_i^0]\right\}_{i=1}^T \in \mathbb{R}^{T \times \left(\frac{HW}{P^2}+M\right) \times C},
\end{equation}
where $m_i^0$ indicates the copycat of messenger tokens $m^0$. The concatenated joint tokens $\left\{[f_i^0, m_i^0]\right\}_{i=1}^T$ are taken as inputs for our MsgShifT.

Our transformer architecture consists of $N_S = 4$ stages and each stage has the same architecture as in Fig.~\ref{fig:overall_arch} (left). The multi-head self attention ($\mathrm{MHSA}$) and feed forward network ($\mathrm{FFN}$) act on the concatenated joint tokens in a per-frame manner:
\begin{equation}
\left\{[f_i^l, m_i^l]\right\}_{i=1}^T = \left\{\mathrm{FFN}^{l}\left(\mathrm{MHSA}^{l}\left([f_i^{l-1}, m_i^{l-1}]\right)\right)\right\}_{i=1}^T.
\end{equation}
Next, a messenger shift manipulation performs temporal information exchange across video frames.

In short, the messenger shift mechanism takes temporal messenger tokens $\left\{m_i^l\right\}_{i=1}^T$ as inputs and builds temporal context modeling by shifting messenger tokens along the temporal axis.
Fig.~\ref{fig:messenger_shift} gives a detailed illustration. First, messenger tokens are divided into $G=4$ groups and shifted along the temporal axis with different time steps ($S=1$ or $2$) and direction (forward or backward).
With various time steps and directions, messenger tokens are able to achieve temporal context exchange with both past and future frames.
Moreover, for every two messenger shift operations, we apply an inverse operation to the second one, which implies the messenger tokens will be shifted back to their original corresponding frames after two contiguous messenger shift manipulations.
This design aims to maintain a stable temporal receptive field as the network goes deeper.

After the above process, the messenger tokens and patch tokens go through one of four stages, and the output tokens are reshaped to feature maps $\left\{F_i^{1}\right\}_{i=1}^T$ which is $\frac{1}{4}$ smaller than the original image. In the same way, using the output messenger tokens and patch tokens of prior stage as inputs, we obtain the following pyramid feature maps $\left\{F_i^{2}\right\}_{i=1}^T$, $\left\{F_i^{3}\right\}_{i=1}^T$ and $\left\{F_i^{4}\right\}_{i=1}^T$, whose strides are $8$, $16$ and $32$ pixels with respect to the input image. The pyramid feature maps $\left\{F^1_i, F^2_i, F^3_i, F^4_i\right\}_{i=1}^T$ will be used to predict video instances in the head network.

MsgShifT performs early temporal fusion in the backbone network, while the previous transformer-based VIS approaches~\cite{vistr, querytrack, tcis, ifc} only perform temporal feature fusion using transformer encoders after image-level feature extraction. It is almost parameter-free, friendly to image-level pre-training models, and brings negligible computation costs. The messenger tokens are randomly initialized and the shift manipulation has no parameter, so this module is insensitive to the pre-training process, which will be further discussed in the experiments in Tab.~\ref{tab:shift_vs_attention}.

\subsection{Spatiotemporal Query Interaction Head}

MsgShifT achieves frame-level spatiotemporal context modeling. Meanwhile, in the VIS head network, our method still emphasizes temporally efficient spatiotemporal context modeling, but at the instance level. To this end, we propose a spatiotemporal query interaction (STQI) head network based on the recent SoTA query-based image-level instance segmentation method, \ie, QueryInst~\cite{queryinst}.

As shown in Fig.~\ref{fig:overall_arch} (right), our head network contains $N_H = 6$ STQI heads and takes a fixed-length instance queries $Q \in \mathbb{R}^{N_q \times C}$ along with pyramid features extracted by MsgShifT $\left\{F^1_i, F^2_i, F^3_i, F^4_i\right\}_{i=1}^T$ as inputs, and generates $N_q \times T$ instance predictions. $N_q$ and $C$ denotes the numbers and the channel dimensions of instance query respectively.
Instance queries $Q$ are randomly initialized and optimized during training. Additionally, our VIS network also contains a set of proposal boxes $B \in \mathbb{R}^{N_q \times 4}$ as prior proposals, for more details about this, we refer readers to QueryInst~\cite{queryinst}.

Instance queries are firstly copied by $T$ times to each frame. Two successive and parameter-shared $\mathrm{MHSA}$ modules act on instance queries along spatial and temporal dimensions:
\begin{equation}
\hat{Q}^{1:T}_{1:N_q} = \left\{\mathrm{MHSA}\left(Q^{i}_{1:N_q}\right)\right\}_{i=1}^{T},
\label{eq:spatial_query_mhsa}
\end{equation}

\begin{equation}
\widetilde{Q}^{1:T}_{1:N_q} = \left\{\mathrm{MHSA}\left(\hat{Q}^{1:T}_{j}\right)\right\}_{j=1}^{N_q}.
\label{eq:temporal_query_mhsa}
\end{equation}
``$1$:$K$'' denotes ranging from $1$ to $K$. Enhanced instance queries $\widetilde{Q}$ are fed into a dynamic convolution module and perform interactions with instance region features. Its output serves as the input queries of the next head.
Finally, task specific heads (\ie, classification head, box head and mask head) predict a sequence of video instances:
\begin{equation}
\left\{\hat{y}_{i}^{t}\right\}_{1:N_q}^{1:T} = \left\{\left(\hat{p}^{t}_{i}(c), \hat{b}_{i}^{t}, \hat{m}_{i}^{t}\right)\right\}_{1:N_q}^{1:T},
\end{equation}
where $\hat{p}(c)$, $\hat{b}$ and $\hat{m}$ denotes predicted confidence scores, bounding boxes and instance foreground masks, respectively.

The advantages of STQI mainly stem from the minimum modifications on the still-image instance prediction head in~\cite{queryinst}. STQI achieves highly efficient temporal context modeling at instance-level by a parameter-shared $\mathrm{MHSA}$ (Eq.~\ref{eq:spatial_query_mhsa} and Eq.~\ref{eq:temporal_query_mhsa}) while does not involve extra parameters.

\begin{table*}
\begin{center}
\setlength{\tabcolsep}{12.3 pt}
\resizebox{\linewidth}{!}{ 
\begin{small}
\begin{tabular}{llc|c|c|c|c|c|c}
\hline

\hline
\rowcolor{mygray}

Method & Backbone & MST & FPS & AP & AP$_{\mathtt{50}}$ & AP$_{\mathtt{75}}$ & AR$_{\mathtt{1}}$ & AR$_{\mathtt{10}}$ \\
\hline
\hline
MaskTrack R-CNN~\cite{vis} & ResNet-$50$ & & $32.8$ & $30.3$ & $51.1$ & $32.6$ & $31.0$ & $35.5$ \\
MaskTrack R-CNN~\cite{vis} & ResNet-$101$ & & $28.6$ & $31.9$ & $53.7$ & $32.3$ & $32.5$ & $37.7$ \\
SipMask~\cite{sipmask} & ResNet-$50$ & \checkmark & $34.1$ & $33.7$ & $54.1$ & $35.8$ & $35.4$ & $40.1$ \\
SG-Net~\cite{sgnet} & ResNet-$50$ & & $-$ & $34.8$ & $56.1$ & $36.8$ & $35.8$ & $40.8$ \\
SG-Net~\cite{sgnet} & ResNet-$101$ & & $-$ & $36.3$ & $57.1$ & $39.6$ & $35.9$ & $43.0$ \\
CrossVIS~\cite{crossvis} & ResNet-$50$ & \checkmark & $39.8$ & $36.3$ & $56.8$ & $38.9$ & $35.6$ & $40.7$ \\
CrossVIS~\cite{crossvis} & ResNet-$101$ & \checkmark & $35.6$ & $36.6$ & $57.3$ & $39.7$ & $36.0$ & $42.0$ \\
\hline
STEm-Seg~\cite{stemseg} & ResNet-$50$ & \checkmark & $3.0$ & $30.6$ & $50.7$ & $33.5$ & $31.6$ & $37.1$ \\
STEm-Seg~\cite{stemseg} & ResNet-$101$ & \checkmark & $-$ & $34.6$ & $55.8$ & $37.9$ & $34.4$ & $41.6$ \\
MaskProp~\cite{maskprop} & ResNet-$50$ & \checkmark & $-$ & $40.0$ & $-$ & $42.9$ & $-$ & $-$ \\
MaskProp~\cite{maskprop} & ResNet-$101$ & \checkmark & $-$ & $42.5$ & $-$ & $45.6$ & $-$ & $-$ \\
MaskProp~\cite{maskprop} & STSN-X-$101$ & \checkmark & $-$ & $\mathbf{46.6}$ & $-$ & $51.2$ & $44.0$ & $52.6$ \\
SeqMask R-CNN~\cite{seqmaskrcnn} & ResNet-$50$ & & $-$ & $40.4$ & $63.0$ & $43.8$ & $41.1$ & $49.7$ \\
SeqMask R-CNN~\cite{seqmaskrcnn} & ResNet-$101$ & & $-$ & $43.8$ & $65.5$ & $47.4$ & $43.0$ & $53.2$ \\
\hline
VisTR~\cite{vistr} & ResNet-$50$ & & $51.1$ & $36.2$ & $59.8$ & $36.9$ & $37.2$ & $42.4$ \\
VisTR~\cite{vistr} & ResNet-$101$ & & $43.5$ & $40.1$ & $64.0$ & $45.0$ & $38.3$ & $44.9$ \\
EfficientVIS~\cite{efficientvis} & ResNet-$50$ & \checkmark & $36.0$ & $37.9$ & $59.7$ & $43.0$ & $40.3$ & $46.6$ \\
EfficientVIS~\cite{efficientvis} & ResNet-$101$ & \checkmark & $32.0$ & $39.8$ & $61.8$ & $44.7$ & $42.1$ & $49.8$ \\
IFC~\cite{ifc} & ResNet-$50$ & \checkmark & $107.1$ & $41.2$ & $65.1$ & $44.6$ & $42.3$ & $49.6$ \\
IFC~\cite{ifc} & ResNet-$101$ & \checkmark & $89.4$ & $42.6$ & $66.6$ & $46.3$ & $43.5$ & $51.4$ \\
\hline
TeViT (ours) & MsgShifT & & $68.9$ & $45.9$ & $69.1$ & $50.4$ & $44.0$ & $53.4$ \\
TeViT (ours) & MsgShifT & \checkmark & $68.9$ & $\mathbf{46.6}$ & $\mathbf{71.3}$ & $\mathbf{51.6}$ & $\mathbf{44.9}$ & $\mathbf{54.3}$ \\
\hline

\hline
\end{tabular}
\end{small}
}
\caption{Comparisons on YouTube-VIS-$2019$ dataset~\cite{vis}. ``\checkmark" under ``MST" indicates using multi-scale training strategy,
The FPS is measured with a single TESLA V$100$ GPU.
All methods in the figure are organized into four groups. According to their basic architectures, the first two groups of methods are built upon CNN architecture, while the last two are transformer-based. According to their training inference paradigms, the first group follows the online and track-by-detect fashion, while the rest all follow offline and sequence-in-sequence-out paradigm.
}
\label{tab:vis2019}
\end{center}
\end{table*}

\subsection{Matching and Loss Function}

The loss function is motivated by \cite{detr}. We first compute the one-to-one assignment between predicted instances and ground-truth annotations.
The ground-truth annotations are denoted as follows:
\begin{equation}
\left\{y_{j}^{t}\right\}_{1:N_{gt}}^{1:T} = \left\{\left(c_{j}^t, b_{j}^t, m_{j}^t\right)\right\}_{1:N_{gt}}^{1:T},
\end{equation}
in which $N_{gt}$ indicates the number of ground-truth video instances, $c$, $b$ and $m$ indicates the category, bounding box and mask respectively.
We then perform sequence-level bipartite matching between predictions and annotations by Hungarian algorithm~\cite{hungarian}. The cost matrix with size of $N_q \times N_{gt}$ between each predicted video instance and each annotation is defined as follows.
\begin{equation}
\begin{aligned}
\mathcal{L}_{Hung}(\hat{y}^{1:T}_i, {y}^{1:T}_j) & = \lambda_{cls} \cdot \mathcal{L}_{cls}(\hat{p}_{i}^{1:T}(c), p_j^{1:T}) \\
& +\lambda_{L1} \cdot \mathcal{L}_{L1}(\hat{b}_{i}^{1:T}, b_{j}^{1:T}) \\ 
& +\lambda_{giou} \cdot \mathcal{L}_{giou}(\hat{b}_{i}^{1:T}, b_{j}^{1:T}),
\end{aligned}
\end{equation}
where $\mathcal{L}_{cls}$ indicates the focal loss~\cite{focalloss} for classification, $\mathcal{L}_{L1}$ and $\mathcal{L}_{giou}$ indicates the L$1$ loss and GIoU loss~\cite{giou} respectively. $\lambda_{cls}, \lambda_{L1}, \lambda_{giou} \in \mathbb{R}$ are hyper-parameters which we simply follow \cite{queryinst, sparsercnn, defdetr}. Besides we use dice coefficient~\cite{diceloss} to optimize mask predictions. For more details, please refer to \cite{queryinst}.

\subsection{Online and Offline Inference}
\label{sec:online_and_offline_inference}

Our method is flexible for both offline and online inference. Under the offline scenario, our TeViT takes the \textit{whole} video clips as inputs and then outputs all possible video instances with a single run. No post-tracking process is needed. When it comes to the near online \cite{stemseg} scenario, an entire video is split into several overlapping segments. TeViT takes clips in time order and generates predictions. A rule-based post-tracking procedure is applied to linking instances across different video clips. For instances from two overlapping video clips, we first compute the similarity score between each instance, and then a Hungarian matcher gives the assignment according to the similarity matrix. The similarity score is defined as a combination of box IoU and mask IoU.

\begin{table*}
\centering
\begin{minipage}[t]{0.49\linewidth}
\centering
\renewcommand\arraystretch{1.17}
\renewcommand\tabcolsep{3.5pt}
\begin{small}
\begin{tabular}{l|ccccc}
\hline

\hline
\rowcolor{mygray}
Methods & AP & AP$_{\mathtt{50}}$ & AP$_{\mathtt{75}}$ & AR$_{\mathtt{1}}$ & AR$_{\mathtt{10}}$ \\
\hline
\hline
MaskTrack R-CNN$^{\dagger}$~\cite{vis, crossvis} & $28.6$ & $48.9$ & $29.6$ & $26.5$ & $33.8$ \\
SipMask$^{\dagger}$~\cite{sipmask, crossvis} & $31.7$ & $52.5$ & $34.0$ & $30.8$ & $37.8$ \\
CrossVIS~\cite{crossvis} & $34.2$ & $54.4$ & $37.9$ & $30.4$ & $38.2$ \\
IFC~\cite{ifc} & $35.2$ & $57.2$ & $37.5$ & $-$ & $-$ \\
TeViT & $37.9$ & $61.2$ & $42.1$ & $35.1$ & $44.6$ \\
\hline

\hline
\end{tabular}
\end{small}
\caption{Comparisons with previous VIS methods on YouTube-VIS-$2021$ datasets. Methods with superscript ``$\dagger$" are reported in~\cite{crossvis}.
}
\label{tab:vis2021}
\end{minipage}
\hfill
\begin{minipage}[t]{0.49\linewidth}
\centering
\renewcommand\arraystretch{1}
\renewcommand\tabcolsep{3.5pt}
\begin{small}
\begin{tabular}{l|ccccc}
\hline

\hline
\rowcolor{mygray}
Methods 
& AP 
& AP$_{\mathtt{50}}$ 
& AP$_{\mathtt{75}}$ 
& AR$_{\mathtt{1}}$ 
& AR$_{\mathtt{10}}$ 
\\
\hline
\hline
SipMask$^{\dagger}$~\cite{sipmask, crossvis} & $10.3$ & $25.4$ & $7.8$ & $7.9$ & $15.8$ \\
MaskTrack R-CNN$^{\dagger}$~\cite{vis, crossvis} & $10.9$ & $26.0$ & $8.1$ & $8.3$ & $15.2$ \\
STEm-Seg$^{\ddagger}$~\cite{stemseg, ovis} & $13.8$ & $32.1$ & $11.9$ & $9.1$ & $20.0$ \\
CrossVIS~\cite{crossvis} & $14.9$ & $32.7$ & $12.1$ & $10.3$ & $19.8$ \\
CMaskTrack R-CNN$^{\ddagger}$~\cite{ovis} & $15.4$ & $33.9$ & $13.1$ & $9.3$ & $20.0$ \\
TeViT & $17.4$ & $34.9$ & $15.0$ & $11.2$ & $21.8$ \\
\hline

\hline
\end{tabular}
\end{small}
\caption{Comparisons on OVIS dataset. Methods with superscript ``$\dagger$" and ``$\ddagger$"  are reported in~\cite{crossvis} and~\cite{ovis} respectively.
}
\label{tab:ovis}
\end{minipage}
\end{table*}

\section{Experiments}

\subsection{Datasets and Evaluation Metrics}
We evaluate TeViT on three challenging video instance segmentation benchmarks, \ie, YouTube-VIS-$2019$~\cite{vis}, YouTube-VIS-$2021$~\cite{vis2021}, and OVIS~\cite{ovis}.
\textbf{YouTube-VIS-$\mathbf{2019}$} is the first dataset that focuses on the VIS problem. It contains $40$ common object categories, $4,883$ unique video instances and about $131k$ high-quality instance-level annotations.
\textbf{YouTube-VIS-$\mathbf{2021}$} dataset is the new version of YouTube-VIS-$2019$ with $1.5\times$ more video frames and $2\times$ more annotations.
\textbf{OVIS} dataset aims to explore the VIS problem under high-occlusion scenarios. It consists of $296k$ high-quality instance masks and $5.80$ instances per video from $25$ semantic categories.
Following previous works, we report the performance on the validation set for all three datasets.
We follow the standard VIS evaluation metrics defined in~\cite{vis}.

\subsection{Implementation Details}
TeViT is built upon the $\mathtt{mmdetection}$ toolbox~\cite{mmdetection}.
Unless otherwise noted, hyper-parameters follow the settings of QueryInst~\cite{queryinst}.
We use $Nq=100$ video instance queries as \cite{ifc, queryinst}.
Due to the temporal efficient designs in TeViT, we do not need to create pseudo video data, \eg \cite{stemseg, ifc}, to train the temporal modeling parameters, instead, we first train a transformer-based QueryInst for image-level instance segmentation on the COCO dataset~\cite{mscoco} and then initialize TeViT with the COCO pre-trained QueryInst weights. Besides, we provide a MindSpore \cite{mindspore} implementation of TeViT.

When training on the VIS datasets, we use the AdamW~\cite{adam} optimizer with an initial learning rate of $0.00025$, and a weight decay of $0.0001$. Especially, the backbone learning rate is slightly lower with a multiplier set to $0.1$. We also apply gradient clipping with a maximal gradient norm of $0.1$. TeViT is trained with a batch size of $16$ and a clip length of $T=5$. The total training process contains $12$ epochs, and the learning rate is decreased by $10$ at the $8$-th and $11$-th epoch respectively.
For example, our TeViT can be trained in about $4$ hours with $8$ Tesla V$100$ GPUs on YouTube-VIS-$2019$, which is much faster than previous transformer-based method (\ie, VisTR \cite{vistr}).
The number of instance queries $N_q$ is set to $100$ for all experiments. Following \cite{vis}, all input frames are resized to $360 \times 640$ in single-scale experiments. Settings of multi-scale training simply follow \cite{sipmask}. For inference, all frames are resized to $360 \times 640$ regardless of the training setups. During inference, we use $T=36$ for most results and report the near online results in ablation study. For main results, we evaluate our framework on YouTube-VIS-$2019$, YouTube-VIS-$2021$, and OVIS datasets, with PVT-B$1$~\cite{pvtv2} based MsgShifT as backbone.

It's noted that all reported results in main results and ablation studies are average performance from multiple runs (\ie, we choose five different random seeds and run each random seed for three times). The standard deviations (\ie, $\sigma_{\mathrm{AP}}$ in following tables) are calculated in the same way.

\subsection{Main Results}

\noindent\textbf{Main results on YouTube-VIS-2019 dataset.} We compare our TeViT to state-of-the-art methods on YouTube-VIS-$2019$ dataset in Tab.~\ref{tab:vis2019}. The longest video in YouTube-VIS-$2021$ dataset only contains $36$ frames, so that our TeViT executes fully offline inference on this dataset. Without bells and whistles, our TeViT achieves $45.9$ AP when using a single-scale training strategy and outperforms the previous state-of-the-art methods by a large margin. Multi-scale training strategy further boosts the performance to $46.6$ AP. Meanwhile, our method also achieves competitive inference speed. With about $10$ AP higher, our method is still faster than VisTR.

\noindent\textbf{Main results on YouTube-VIS-2021 dataset.}
Tab.~\ref{tab:vis2021} shows the final results of several VIS methods and ours on YouTube-VIS-$2021$ dataset. Due to the video length in YouTube-VIS-$2021$ is longer than our inference clip length ($T=36$), TeViT performs near online tracking described in Sec.~\ref{sec:online_and_offline_inference} on this dataset. TeViT obtains $37.9$ AP, outperforming the previous state-of-the-art method by $2.7$ AP.

\noindent\textbf{Main results on OVIS dataset.}
The results on the OVIS dataset are shown in Tab.~\ref{tab:ovis}. Our method also performs near online inference on OVIS dataset. TeViT achieves a relatively higher performance of $17.4$ AP on the $\mathtt{val}$ split, surpassing previous state-of-the-art methods. Compared to CMaskTrack R-CNN~\cite{ovis} which presents an elaborate-designed feature calibration plug-in to alleviate occlusion, our TeViT still gains $2.0$ AP improvement, which shows that our temporal context modeling designs are helpful to segment occluded instances.
\begin{table*}
\centering
\begin{minipage}[t]{0.49\linewidth}
\centering
\renewcommand\arraystretch{1}
 \renewcommand\tabcolsep{1.3pt}
\begin{small}
\begin{tabular}{cc|cccccc}
\hline

\hline
\rowcolor{mygray}
MSM & STQI & GFLOPs & AP $\pm \sigma_{\text{AP}}$ & AP$_{\mathtt{50}}$ & AP$_{\mathtt{75}}$ & AR$_{\mathtt{1}}$ & AR$_{\mathtt{10}}$ \\
\hline
\hline
& & $81.97$ & $42.5 \pm 0.47$ & $67.6$ & $44.0$ & $43.0$ & $52.7$ \\
\checkmark & & $82.19$ & $43.1_{\uparrow(+0.6)} \pm 0.71$ & $67.2$ & $47.8$ & $43.5$ & $52.4$ \\
& \checkmark & $81.97$ & $45.2_{\uparrow(+2.7)} \pm 0.85$ & $68.9$ & $50.2$ & $\mathbf{44.0}$ & $53.0$ \\
\checkmark & \checkmark & $82.19$ & $\mathbf{45.9}_{\uparrow(+3.4)} \pm 0.58$ & $\mathbf{69.1}$ & $\mathbf{50.4}$ & $\mathbf{44.0}$ & $\mathbf{53.4}$ \\
\hline

\hline
\end{tabular}
\end{small}
\caption{Component-wise analysis on TeViT. MSM denotes the messenger shift mechanism and STQI denotes spatiotemporal query interaction. Without applying STQI implies only one $\mathrm{MHSA}$ is performed for query interaction within each frame (excluding Eq.~\ref{eq:temporal_query_mhsa}).}
\label{tab:component_wise}
\end{minipage}
\hfill
\begin{minipage}[t]{0.49\linewidth}
\centering
\renewcommand\arraystretch{1.25}
 \renewcommand\tabcolsep{3.1pt}
\begin{small}
\begin{tabular}{c|ccccc}
\hline

\hline
\rowcolor{mygray}
Interaction & AP & AP$_{\mathtt{50}}$ & AP$_{\mathtt{75}}$ & AR$_{\mathtt{1}}$ & AR$_{\mathtt{10}}$ 
\\
\hline
\hline
Spatial Only~\cite{queryinst}& $43.1$ & $67.2$ & $47.8$ & $43.5$ & $52.4$ \\
Fused Space-Time~\cite{vistr} & $43.9_{\uparrow{(+0.8)}}$ & $\mathbf{69.5}$ & $48.4$ & $42.9$ & $52.0$ \\
Ours & $\mathbf{45.9}_{\uparrow{(+2.7)}}$ & $69.1$ & $\mathbf{50.4}$ & $\mathbf{44.0}$ & $\mathbf{53.4}$ \\
\hline

\hline
\end{tabular}
\end{small}
\caption{Variants of spatiotemporal query interaction. ``Spatial Only'' denotes the image-level instance segmentation heads in \cite{queryinst}, ``Fused Space-Time'' denotes applying $\mathrm{MHSA}$ to all video instance queries at a single run, which is the same as in \cite{vistr}.}
\label{tab:variants_query_interaction}
\end{minipage}
\end{table*}

\begin{table*}
\centering
\begin{minipage}[t]{0.33\linewidth}
\centering
\renewcommand\arraystretch{1.25}
\renewcommand\tabcolsep{5pt}
\begin{small}
\scalebox{0.9}{
\begin{tabular}{c|c|ccc}
\hline

\hline
\rowcolor{mygray}
Manip. & AP $\pm \ \sigma_{\text{AP}}$ & AP$_{\mathtt{50}}$ & AP$_{\mathtt{75}}$ 
\\
\hline
\hline
$\mathrm{None}$ & $45.2 \pm 0.85$ & $68.9$ & $50.2$ \\
$\mathrm{MHSA}+\mathrm{FFN}$ & $44.5 \pm 1.07$ & $69.2$ & $49.3$ \\
$\mathrm{Shift}$ & $\mathbf{45.9 \pm 0.58}$ & $\mathbf{69.1}$ & $\mathbf{50.4}$ \\
\hline

\hline
\end{tabular}}
\caption{Study of the manipulations upon messenger tokens. Our method obtains the highest AP and a relatively stable performance ($\sigma_{\text{AP}}$) among all settings.}
\label{tab:shift_vs_attention}
\end{small}
\end{minipage}
\hfill
\begin{minipage}[t]{0.3\linewidth}
\centering
\renewcommand\arraystretch{1}
 \renewcommand\tabcolsep{5pt}
\begin{small}
\scalebox{0.9}{
\begin{tabular}{c|ccccc}
\hline

\hline
\rowcolor{mygray}
Manip. & AP & AP$_{\mathtt{50}}$ & AP$_{\mathtt{75}}$
\\
\hline
\hline
$\mathrm{None}$ & $45.2$ & $68.9$ & $50.2$ \\
$\mathrm{Conv}$ & $41.8$ & $63.7$ & $45.1$ \\
$\mathrm{MHSA}+\mathrm{FFN}$ & $43.1$ & $67.2$ & $49.1$ \\
$\mathrm{Msg \ Shift}$ & $\mathbf{45.9}$ & $\mathbf{69.1}$ & $\mathbf{50.4}$ \\
\hline

\hline
\end{tabular}}
\caption{Study of frame-level feature aggregation. Compared to other frame-level feature manipulations, our messenger shift (Row $4$) obtains the best results.}
\label{tab:frame_level_agg}
\end{small}
\end{minipage}
\hfill
\begin{minipage}[t]{0.33\linewidth}
\centering
\renewcommand\arraystretch{1.25}
\renewcommand\tabcolsep{6pt}
\begin{small}
\scalebox{0.9}{
\begin{tabular}{c|ccccc}
\hline

\hline
\rowcolor{mygray}
 M & AP & AP$_{\mathtt{50}}$ & AP$_{\mathtt{75}}$ & AR$_{\mathtt{1}}$  & AR$_{\mathtt{10}}$  \\
\hline
\hline
$8$ & $45.3$ & $69.0$ & $48.9$ & $\mathbf{44.5}$ & $52.4$ \\
$16$ & $45.4$ & $\mathbf{70.3}$ & $49.9$ & $44.0$ & $51.7$ \\
$32$ & $\mathbf{45.9}$ & $69.1$ & $\mathbf{50.4}$ & $44.0$ & $\mathbf{53.1}$ \\
\hline

\hline
\end{tabular}}
\caption{Impact of messenger token numbers. M indicates the number of messenger tokens. We increase M from $8$ to $32$ and observe the effects on final performance.}
\label{tab:messenger_token_number}
\end{small}
\end{minipage}
\end{table*}

\subsection{Ablation Study}

\noindent\textbf{Effect of frame-level \& instance-level temporal context modeling.}
We investigate the effects of messenger shift mechanism and spatiotemporal query interaction individually and simultaneously in Tab.~\ref{tab:component_wise}.
Using messenger shift mechanism and spatiotemporal query interaction individually brings $0.6$ and $2.7$ AP improvements respectively.
The results show that both frame-level and instance-level temporal context modeling can obviously improve VIS performance. In addition, the instance-level one brings more significant performance gain. The two designs together brings $3.4$ ($ > 0.6 + 2.7$) AP improvements over a high-performance baseline. Besides the remarkable performance improvements, our designs only bring $0.27\%$ computation overhead on our baseline ($82.19$ GFLOPs \vs $81.97$ GFLOPs), which demonstrates our design is very efficient.

\noindent\textbf{Variants of spatiotemporal query interaction.}
In Tab.~\ref{tab:variants_query_interaction}, we investigate the effectiveness of our spatiotemporal query interaction comparing to its variants. A naive query interaction method in~\cite{queryinst} without using temporal interaction, denoted as ``Spatial Only'', serves as a baseline. ``Fused Space-Time'' in the Row~$2$ denotes fusing video instance queries together and performing spatial and temporal interaction  within a single $\mathrm{MHSA}$, which is the same as in \cite{vistr}. As the results show: (1) Our spatiotemporal query interaction achieves the best performance among three variants.
(2) Compared to the spatial-only query interaction, joint spatiotemporal query interaction brings only $0.8$ AP improvements. However, our method achieves $2.7$ AP gains. We argue this is because the one-to-one corresponding between instance queries are misaligned in the joint spatiotemporal attention, while ours is not.

\noindent\textbf{Different messenger token manipulation methods.}
We compare our messenger shift mechanism with two other optional manipulations in Tab.~\ref{tab:shift_vs_attention}.
$\mathrm{None}$ indicates there are no extra manipulations conducted on messenger tokens, thus no temporal information is employed.
$\mathrm{MHSA}+\mathrm{FFN}$ stands for the same operation in \cite{ifc} which performs extra $\mathrm{MHSA}$ and $\mathrm{FFN}$ on messenger tokens, and $\mathrm{Shift}$ denotes our messenger shift mechanism.
Different from \cite{ifc}, we do not conduct any extra pre-training process so that both the messenger tokens and $\mathrm{MHSA}$ layers with $\mathrm{FFN}$ are randomly initialized and trained from scratch.
As results have shown: (1) Our method achieves the best AP and outperforms $\mathrm{None}$ by $0.7$ AP and $\mathrm{MHSA}+\mathrm{FFN}$ by $1.4$ AP. 
(2) Compared to conducting $\mathrm{MHSA}+\mathrm{FFN}$ upon messenger tokens, our method obtains a more stable performance. $\mathrm{MHSA}+\mathrm{FFN}$ shows more fluctuating final results (see $\sigma_{\text{AP}}$ in Tab.~\ref{tab:shift_vs_attention}) while ours is much more stable.
The results confirm that the randomized $\mathrm{MHSA}+\mathrm{FFN}$ is unable to capture temporal context while our parameter-free shift operation works.

\noindent\textbf{Different manipulations on frame-level feature aggregation.}
We also compare our messenger shift mechanism with other optional frame-level feature aggregation manipulations in Tab.~\ref{tab:frame_level_agg}.
$\mathrm{None}$ indicates no manipulation is conducted to aggregate frame-level temporal features.
$\mathrm{Conv.}$ and $\mathrm{MHSA}+\mathrm{FFN}$ denotes using newly introduced convolution layers or transformer layers to achieve temporal feature aggregation.
As results have shown, our messenger shift mechanism achieves the best performance (\ie, $45.9$ AP) compared to its all counterparts.
Meanwhile, we observe apparent performance decrease by using newly introduced $\mathrm{Conv.}$ or $\mathrm{MHSA}+\mathrm{FFN}$ as aggregation layers.
We infer such a performance decrease comes from the enormous newly introduced parameters, while our messenger shift mechanism eliminates this performance decrease by the nearly parameter-free design.

\noindent\textbf{Number of messenger tokens.}
In Tab.~\ref{tab:messenger_token_number}, we test our method with number of messenger tokens increases from $8$ to $32$.
Compared to less messenger tokens ($M=8, 16$), more messenger tokens ($M=32$) achieves better results.
Unless specified, our experiments are conducted with $32$ messenger tokens.

\begin{table*}
\centering
\begin{minipage}[t]{0.49\linewidth}
\centering
\renewcommand\arraystretch{1.4}
\renewcommand\tabcolsep{12.2pt}
\begin{small}
\begin{tabular}{c|ccccc}
\hline

\hline
\rowcolor{mygray}
T & AP & AP$_{\mathtt{50}}$ & AP$_{\mathtt{75}}$ & AR$_{\mathtt{1}}$ & AR$_{\mathtt{10}}$ 
\\
\hline
\hline
$2$ & $41.1$ & $64.8$ & $44.3$ & $41.2$ & $50.2$ \\
$3$ & $44.3$ & $69.2$ & $49.3$ & $43.7$ & $52.1$ \\
$5$ & $45.9$ & $68.9$ & $50.2$ & $44.0$ & $53.0$ \\
$7$ & $46.3$ & $71.9$ & $51.6$ & $44.0$ & $53.4$ \\
\hline

\hline
\end{tabular}
\end{small}
\caption{Effect of training clip length on AP. ``T'' indicates the number of frames for each video clip during training.}
\label{tab:train_clip_length}
\end{minipage}
\hfill
\begin{minipage}[t]{0.49\linewidth}
\centering
\renewcommand\arraystretch{1}
\renewcommand\tabcolsep{9.5pt}
\begin{small}
\begin{tabular}{cc|ccccc}
\hline

\hline
\rowcolor{mygray}
T & S & AP & AP$_{\mathtt{50}}$ & AP$_{\mathtt{75}}$ & AR$_{\mathtt{1}}$ & AR$_{\mathtt{10}}$ 
\\
\hline
\hline
$5$ & $1$ & $42.1$ & $66.8$ & $46.7$ & $41.5$ & $51.3$ \\
$5$ & $3$ & $41.7$ & $63.2$ & $46.1$ & $41.8$ & $50.8$ \\
$10$ & $5$ & $44.1$ & $66.4$ & $48.3$ & $42.9$ & $51.9$ \\
$15$ & $8$ & $44.7$ & $67.0$ & $48.7$ & $43.4$ & $52.5$ \\
$20$ & $10$ & $46.0$ & $67.5$ & $50.0$ & $43.7$ & $53.1$ \\
$36$ & $18$ & $45.9$ & $68.9$ & $50.2$ & $44.0$ & $53.0$ \\
\hline

\hline
\end{tabular}
\end{small}
\caption{Study the impact of clip length and stride during inference phase. ``T'' and ``S'' indicates the clip length and overlapping stride respectively.}
\label{tab:inference_clip_length}
\end{minipage}
\end{table*}

\begin{table*}
\centering
\begin{minipage}[t]{0.49\linewidth}
\centering
\renewcommand\arraystretch{1.0}
\renewcommand\tabcolsep{4.1pt}
\begin{small}
\begin{tabular}{lcc|ccccc}
\hline

\hline
\rowcolor{mygray}
Method & MST & FPS & AP & AP$_{\mathtt{50}}$ & AP$_{\mathtt{75}}$ & AR$_{\mathtt{1}}$ & AR$_{\mathtt{10}}$ 
\\
\hline
\hline
VisTR~\cite{vistr} & & $51.1$ & $36.2$ & $59.8$ & $36.9$ & $37.2$ & $42.4$ \\
IFC~\cite{ifc} & \checkmark & $107.1$ & $41.2$ & $65.1$ & $44.6$ & $42.3$ & $49.6$ \\
\multicolumn{2}{l}{Ours w/o. Eq.~\ref{eq:temporal_query_mhsa}} & $78.1$ & $36.8$ & $78.3$ & $38.8$ & $38.3$ & $46.2$ \\
Ours & & $76.8$ & $41.7$ & $67.8$ & $44.8$ & $41.3$ & $49.9$ \\
Ours & \checkmark & $76.8$ & $42.3$ & $67.6$ & $44.0$ & $43.0$ & $52.7$ \\
\hline

\hline
\end{tabular}
\end{small}
\caption{Comparisons with ResNet-$50$ as backbone.}
\label{tab:resnet_backbone}
\end{minipage}
\hfill
\begin{minipage}[t]{0.49\linewidth}
\centering
\renewcommand\arraystretch{1.5}
\renewcommand\tabcolsep{4.2pt}
\begin{small}
\begin{tabular}{cc|ccccc}
\hline

\hline
\rowcolor{mygray} Train & Inference & AP & AP$_{\mathtt{50}}$ & AP$_{\mathtt{75}}$ & AR$_{\mathtt{1}}$ & AR$_{\mathtt{10}}$ 
\\
\hline
\hline
learnable & learned & $45.9$ & $68.9$ & $50.2$ & $44.0$ & $53.0$ \\
learnable & zero & $45.5_{\downarrow(-0.4)}$ & $68.5$ & $49.9$ & $43.8$ & $52.5$ \\
learnable & random & $45.6_{\downarrow(-0.3)}$ & $68.2$ & $49.7$ & $43.7$ & $52.4$ \\
\hline

\hline
\end{tabular}
\end{small}
\caption{Revisiting messenger tokens in inference phase.}
\label{tab:understanding_messenger_token}
\end{minipage}
\end{table*}

\noindent\textbf{Training and inference clip length.}
We also investigate the effects of clip length in both the training and testing phase.
From Tab.~\ref{tab:train_clip_length}, we find that: (1) Our method shows great tolerance to short length of training clip. Only trained with $2$ or $3$ frames, our method can effectively learn temporal context and obtains comparable results to previous methods. (2) The performance improvements by increasing the length of the training clip gradually gets saturated. Increasing training clip length from $2$ to $3$ and $3$ to $5$ brings $3.2$ AP and $1.6$ AP gains respectively while increasing training clip length from $5$ to $7$ only obtains slight $0.4$ AP profit. Besides, a longer training clip requires more training computations and memory budgets. To this end, we set the training clip length of our method to $T=5$ as a compromise between performance and training costs.

Tab.~\ref{tab:inference_clip_length} gives the results of TeViT under different inference settings. ``T'' indicates the input clip length during the inference phase, and ``S'' indicates strides. It shows that our TeViT obtains promising performance under various inference setups. Even with $T=5$ and $S=3$, TeViT still achieves $41.7$ AP, which indicates TeViT can serve as a strong baseline for both offline and online video understanding scenarios.

\noindent\textbf{Performance under ResNet backbone.}
We compare our method with previous transformer-based methods using ResNet-$50$~\cite{resnet} as backbone network in Tab.~\ref{tab:resnet_backbone}.
As results show, our method achieves $41.7$ AP on YouTube-VIS-$2019$ dataset with single-scale training (Row~$4$), even outperforming VisTR and IFC with multi-scale training strategy.
The performance goes a step further to $42.3$ AP when the multi-scale strategy is applied (Row~$5$).
It's worth noting that with ResNet-$50$ as backbone network, the messenger tokens, and messenger shift mechanism are unable to proceed, so our method achieves such high performance with only the STQI mechanism.
We also investigate the improvements by our STQI head. Simply taking off the spatial $\mathrm{MHSA}$ in Eq.~\ref{eq:temporal_query_mhsa}, the final performance drops from $41.7$ AP to $36.8$ AP (Row~$3$), demonstrating the effectiveness of our STQI head directly.

\noindent\textbf{Revisiting messenger tokens.}
Inspired by MSG-Transformer~\cite{msgtransformer}, we re-initialize messenger tokens in inference phase and obvious the influence on performance in Tab.~\ref{tab:understanding_messenger_token}.
As the results show, when we re-initialize messenger tokens to zero, the performance merely drops $0.4$ AP (compare Row~$2$ to Row~$1$).
Randomly initialize the messenger token in inference phase leads to a similar performance decrease (Row~$3$).
We think this phenomenon implies that the messenger tokens contain only a few or not specific information in themselves.
On the contrary, they play the role of summarizing frame-level contexts, and exchanging them across adjacent frames.

\section{Conclusion}
\label{sec:conclusion}

In this paper, we provide lightweight and effective solutions to fully exploit temporal context for VIS. Based on existing ViTs and query-based image-level instance segmentation methods, we proposes the TeViT VIS method that contains the messenger shift and spatiotemporal query interaction mechanisms. TeViT performs both frame-level and instance-level temporal feature interactions while only bringing a few parameters and marginal extra computational costs. Experiments on YouTube-VIS-$2019$, YouTube-VIS-$2021$, and OVIS show that TeViT can obtain remarkably better results than previous SoTA methods, \eg, IFC, VisTR, MaskProp, and STEm-Seg. We believe the proposed temporal context modeling mechanisms have great potential to be extended to other video understanding tasks.

\noindent\textbf{Limitations.}
Although the extensive experiments have demonstrated the capacity and efficiency of our TeViT on temporal context modeling, it still suffers effects from occlusion, motion deformation and long time-span videos (\ie, results of TeViT in Tab.~\ref{tab:vis2021} and Tab.~\ref{tab:ovis} are far from satisfying). We leave these promising directions as future work.

\noindent\textbf{Broader impact.}
Although our research does not make direct negative impacts in society, it may be misused by illegal video applications, which could be a potential invasion to human privacy.

\noindent\textbf{Acknowledgement.} This work was in part supported by NSFC (No. 61876212 and No. 61733007) and CAAI-Huawei MindSpore Open Fund.

{\small
\bibliographystyle{ieee_fullname}
\bibliography{egbib}
}

\end{document}